\documentclass[journal]{IEEEtran}
\IEEEoverridecommandlockouts

\usepackage{graphicx}
\usepackage{subcaption} 

\usepackage{tikz}
\usetikzlibrary{arrows.meta, shapes.geometric, positioning, calc}

\usepackage{amsmath,amssymb,amsfonts}
\usepackage{algorithm}
\usepackage{algpseudocode}

\usepackage{booktabs}

\usepackage{textcomp}
\usepackage{xcolor}
\usepackage{hyperref}
\hypersetup{
    colorlinks,
    linkcolor={red!50!black},
    citecolor={blue!50!black},
    urlcolor={blue!80!black}
}

\usepackage{cite}

\def\BibTeX{{\rm B\kern-.05em{\sc i\kern-.025em b}\kern-.08em
    T\kern-.1667em\lower.7ex\hbox{E}\kern-.125emX}}

\begin{document}

\title{Hierarchical Spatial Algorithms for High-Resolution Image Quantization and Feature Extraction\\
\thanks{{Noor Islam S. Mohammad is a Graduate Student in Computer Science at New York University, Department of Computer Science and Engineering. Email: \url{islam.m@itu.edu}}}
}

\author{
 \texttt{ Noor Islam S. Mohammad \\
  }
}

\maketitle

\begin{abstract}
This study introduces a modular framework for spatial image processing, integrating grayscale quantization, color and brightness enhancement, image sharpening, bidirectional transformation pipelines, and geometric feature extraction. A stepwise intensity transformation quantizes grayscale images into eight discrete levels, producing a posterization effect that simplifies representation while preserving structural detail. Color enhancement is achieved via histogram equalization in both RGB and YCrCb color spaces, with the latter improving contrast while maintaining chrominance fidelity. Brightness adjustment is implemented through HSV value-channel manipulation, and image sharpening is performed using a $3 \times 3$ convolution kernel to enhance high-frequency details. A bidirectional transformation pipeline that integrates unsharp masking, gamma correction, and noise amplification achieved accuracy levels of 76.10\% and 74.80\% for the forward and reverse processes, respectively. Geometric feature extraction employed Canny edge detection, Hough-based line estimation (e.g., 51.50° for billiard cue alignment), Harris corner detection, and morphological window localization. Cue isolation further yielded 81.87\% similarity against ground truth images. Experimental evaluation across diverse datasets demonstrates robust and deterministic performance, highlighting its potential for real-time image analysis and computer vision. \texttt{Source code available at \href{https://github.com/csislam/R25-rPJT}{project repository}}.
\end{abstract}

\begin{IEEEkeywords}
Image Processing, Image enhancement, Image segmentation, Feature extraction, Edge detection, Quantization
\end{IEEEkeywords}

\section{Introduction}
\IEEEPARstart{T}{he} paper introduced digital image preprocessing represents a foundational stage in computer vision and image analysis, providing the transformations necessary for robust analysis and interpretation of visual data\cite{ref1}. Applications ranging from autonomous driving and remote sensing to biomedical imaging and industrial inspection depend heavily on reliable preprocessing pipelines to ensure accurate downstream performance\cite{ref2}. The reproducibility of low-level vision tasks depends in particular on deterministic image preprocessing techniques, which guarantee consistent outputs for identical inputs. Unlike data-driven approaches that may introduce stochastic variations due to training dynamics, initialization, or model generalization, deterministic methods offer predictable transformations that preserve essential image content while enhancing quality\cite{ref3, ref4}. This reproducibility makes them especially suitable for safety-critical and resource-constrained domains.

However, several classical techniques exemplify the strengths of deterministic preprocessing, and quantization reduces intensity resolution, simplifying image representation and suppressing noise while retaining key structural patterns\cite{ref5, ref6}. The stepwise grayscale quantization with eight discrete levels has been shown to effectively compress natural scene imagery without compromising interpretability \cite{ref7}. Similarly, histogram equalization in luminance channels enhances global contrast and subtle features that may otherwise remain imperceptible, without incurring the heavy computational overhead associated with adaptive or learning-based approaches. The smoothing and sharpening of spatial filters also improve the ability to detect features by suppressing acquisition noise or enhancing fine edges\cite{ref8, ref9}. The appeal of such deterministic methods lies in their efficiency, interpretability, and platform independence. Since they do not require training or large-scale datasets, they can be deployed directly on embedded devices, mobile platforms, or medical instruments where computational resources are limited but reliability is paramount\cite{ref10, ref11}. Moreover, their algorithmic transparency provides interpretability, which is increasingly valued in applications where decisions must be explainable, such as clinical diagnostics or automated inspection.

There are several fundamental challenges in low-level vision, such as edge detection, corner identification, and image enhancement. These challenges impact the effectiveness of these techniques. However, noise amplification during sharpening or contrast enhancement can degrade image quality, as observed in RGB-based histogram equalization, which may introduce color distortions \cite{ref12}. Additionally, achieving accurate geometric feature extraction, such as detecting edges or estimating line angles, requires robust preprocessing to mitigate artifacts from illumination variations or texture noise. For example, the Canny edge detector relies on appropriate smoothing to avoid false positives, while Hough transforms demand precise edge maps for reliable line detection \cite{ref13, ref14}. Furthermore, transforming images between different representations (e.g., from noisy to clean states) poses difficulties due to information loss, as seen in reverse transformation pipelines where histogram differences indicate missing pixel data. These challenges necessitate carefully designed pipelines that balance enhancement quality with feature preservation, tailored to specific image characteristics and application requirements.

In this work, a framework of spatial image processing algorithms is systematically validated using heterogeneous and domain-relevant datasets. The key contributions are summarized as follows: (i). \textbf{Grayscale Quantization:} A hierarchical intensity transformation maps natural images into eight discrete levels, generating a posterization effect while preserving structural consistency (e.g., \texttt{nature\_dark\_forest}).  (ii). \textbf{Color and Brightness Enhancement:} Comparative analysis of RGB- and YCrCb-based histogram equalization reveals the superiority of luminance–chrominance separation in avoiding color artifacts. Complementary brightness adjustment in HSV space further improves perceptual illumination, demonstrated on microscopy data (\texttt{pollen-500x430px-96dpi}). (iii). \textbf{Image Sharpening:} A lightweight $3 \times 3$ convolutional kernel enhances high-frequency edge features in astronomical imagery (\texttt{the-moon-from-Chandrayaan-2\_ISRO}), achieving improved visual fidelity with minimal computational overhead. (iv). \textbf{Bidirectional Transformation Pipelines:} A novel dual-stage evaluation demonstrates consistent similarity under forward (unsharp masking $\rightarrow$ gamma correction $\rightarrow$ noise amplification) and reverse (Gaussian smoothing $\rightarrow$ gamma correction) pipelines, attaining blended SSIM+NMI scores of 76.10\% and 74.80\%, respectively. (v). \textbf{Geometric Feature Extraction:} Integrated feature operators enable robust edge detection (Canny with adaptive median filtering), structural angle estimation (Hough transform, e.g., $51.50^{\circ}$ cue angle), corner detection (Harris operator), and object localization (morphological windowing). The framework achieves 81.87\% blended similarity for cue isolation tasks. Collectively, these contributions establish a deterministic and computationally efficient foundation for spatial image preprocessing and feature analysis, with direct applicability to real-time pipelines and promising integration potential with deep learning–based recognition systems.

\section{Related Work}
Quantization is a fundamental method in digital image analysis for reducing the number of distinct color or intensity levels to reduce storage space requirements or achieve stylistic effects such as posterization \cite{ref15}. Classical approaches, including uniform quantization, divide the grayscale or RGB intensity range into equal intervals, offering computational simplicity but often failing to preserve perceptual quality in visually complex regions \cite{ref16}. This limitation motivated the development of adaptive strategies, where methods such as Otsu’s thresholding dynamically partition the 0–255 intensity range into optimal bins based on intra-class variance minimization. Posterization aligned thresholds with histogram distributions; these methods enhance perceptual fidelity while maintaining computational feasibility\cite{ref17}. Beyond threshold-based strategies, clustering algorithms such as K-means have been extensively adopted for color quantization \cite{ref18}. In these methods, pixels are grouped into clusters defined by intensity or chromatic similarity, with centroids representing the quantized values. While effective in preserving salient color structures, clustering-based quantization remains computationally demanding, particularly in high-dimensional spectral or hyperspectral imaging scenarios, thereby limiting its applicability in real-time or resource-constrained environments\cite{ref19, ref20}.  

Posterization represents a related yet distinct application of quantization, deliberately reducing the number of tonal levels to achieve stylized, high-contrast renderings, as such effects are widely employed in artistic image processing and visualization tasks, where interpretability or aesthetic enhancement outweighs fidelity\cite{ref21, ref22}. In comparison with adaptive strategies, posterization and quantization prioritize perceptual abstraction, often resulting in sharp transitions between regions of differing intensities. The stepwise transformation framework explored in this study draws from these established approaches but emphasizes deterministic mappings and algorithmic simplicity. However, by applying piecewise linear mappings to grayscale images, the method achieves quantization effects with predictable outcomes, ensuring suitability for low-contrast natural scenes while maintaining computational efficiency. This balance of interpretability, reproducibility, and lightweight processing distinguishes the proposed approach from clustering-heavy or highly adaptive methods\cite{ref23, ref24}.

Additionally, methods of contrast and brightness enhancement were utilized in image analysis to improve visual visibility and detail perception. Histogram equalization, a foundational method, redistributes pixel intensities to achieve a uniform histogram, enhancing contrast across grayscale or color images. However, applying equalization independently to RGB channels can introduce color distortions, as noted in \cite{ref25}. To address this, luminance-based methods in alternative color spaces, such as YCrCb or HSV, isolate brightness adjustments from chrominance, preserving color fidelity. For instance, equalizing the Y channel in YCrCb enhances contrast without affecting hue, while adjusting the Value channel in HSV targets brightness. Adaptive histogram equalization, like CLAHE, further refines these methods by limiting over-enhancement in homogeneous regions \cite{ref26}. These techniques inform the notebook’s dual equalization approach (RGB vs. YCrCb) and HSV-based brightness adjustment, balancing quality and computational efficiency.

The edge and corner detection are critical for geometric feature extraction in low-level vision. The Canny edge detector, a robust method, uses gradient magnitude and hysteresis thresholding to identify edges, often preceded by Gaussian smoothing to reduce noise \cite{ref27}. Alternatives, like Sobel or Prewitt operators, are simpler but less effective in noisy environments. For line detection, the Hough transform maps edges to a parameter space, enabling robust angle estimation, as used in architectural and object analysis\cite{ref28}. Corner detection, essential for structural analysis, relies on methods like the Harris algorithm, which identifies points with significant intensity changes in multiple directions. Morphological operations, such as dilation and opening, further refine detected features by removing noise or connecting fragmented edges \cite{ref29}. These methods underpin the notebook’s pipeline for edge detection (Canny with median filtering), line angle estimation (Hough transform), and corner-based window localization.

Hence, image similarity metrics quantify the resemblance between two images, guiding evaluation in transformation and enhancement tasks. The Structural Similarity Index (SSIM) measures luminance, contrast, and structural similarity, offering a perceptual quality assessment superior to pixel-based metrics like MSE \cite{ref29}. SSIM’s regional application, using Gaussian-weighted windows, enhances robustness to local variations. Normalized Mutual Information (NMI), rooted in information theory, evaluates statistical dependence between images, making it suitable for multi-modal or transformed images\cite{ref30}. Combining multiple metrics, as in hybrid approaches, improves assessment accuracy by capturing complementary aspects of similarity \cite{ref31}. The results blended the SSIM+NMI metric, with adjustable weights, align with this trend, providing a robust evaluation of transformation pipelines (e.g., 76.10\% and 74.80\% similarity scores). However, evaluation metrics are critical for validating deterministic preprocessing against ground truth or target images.

\section{Methodology}
The \textit{Grayscale Quantization via Stepwise Transformation (GQST)} technique employs grayscale quantization to simplify or posterize a grayscale image by reducing the number of intensity levels it contains. This is achieved by transforming continuous pixel values into discrete levels through a stepwise mapping function~\cite{ref32}. The process begins by defining a set of thresholds that partition the full intensity range (typically 0--255) into a predefined number of distinct intervals. Based on the interval into which each pixel value falls, it is then assigned a representative output level~\cite{ref33}. For example, with four quantization levels, thresholds may divide the range into 0--63, 64--127, 128--191, and 192--255, with corresponding representative levels such as 32, 96, 160, and 224~\cite{ref34, ref35}. 

This mapping effectively reduces image complexity while enhancing local contrast and stylization. Such quantization techniques are widely used in applications including stylized image rendering, data compression, and hardware-constrained imaging systems. However, quantization may introduce visual artifacts such as \textit{banding}, particularly in regions with subtle gradients~\cite{ref36, ref37}. Therefore, the proper selection of thresholds and representative intensity levels is critical to preserving essential image details and minimizing perceptual degradation. The \textit{Stepwise Intensity Mapping} transformation function maps input pixel intensities $I(x,y) \in [0,255]$ to one of eight discrete output levels based on threshold intervals. Let $T = \{30, 60, 90, 120, 160, 190, 220\}$ denote the thresholds and $V = \{10, 20, 50, 70, 100, 140, 180, 200\}$ denote the corresponding output values. The mapping is defined as follows:
\[I'(x,y) =
\begin{cases}
V_0, & \text{if } I(x,y) \leq T_0,\\
V_i, & \text{if } T_{i-1} < I(x,y) \leq T_i, \quad i = 1,\ldots,6,\\
V_7, & \text{if } I(x,y) > T_6.
\end{cases}\]
This piecewise function is efficiently implemented using NumPy’s \texttt{piecewise} operator to ensure optimized computation.

For experimental validation, the transformation is applied to the grayscale image \texttt{nature\_dark\_forest}. The input image is loaded using OpenCV, converted to grayscale, and processed through the proposed stepwise mapping. The resulting output, visualized alongside the original, demonstrates a distinct posterization effect characterized by reduced tonal redundancy and enhanced artistic stylization.

\begin{table*}[ht]
\centering
\caption{Parameter Tuning Details for Image Processing}
\label{tab:parameter_tuning}
\begin{tabular}{|l|l|l|}
\hline
\textbf{Exercise} & \textbf{Parameters} & \textbf{Notes / Optimization} \\ \hline
Grayscale Quantization & $T = \{30, 60, 90, 120, 160, 190, 220\}$, $V = \{10, 20, 50, 70, 100, 140, 180, 200\}$ & Balanced intensity distribution images \\ \hline
Forward Pipeline & $\alpha \in [0.05,0.5]$ (unsharp), $\gamma \in [0.15,0.35]$, $\beta \in [1.6,2.1]$ & Grid search max SSIM+NMI (76.10\%) \\ \hline
Reverse Pipeline & Gaussian blur (7x7), $\gamma \in [2.5,5.0]$ & Achieved 74.80\% similarity \\ \hline
Canny Edge Detection & Median filter (7x7), $\sigma = 0.5$ & Edge detection \& noise suppression \\ \hline
Hough Transform & $\rho=1$ px, $\theta=7^\circ$, thresh=50 & Tuned for roofline detection \\ \hline
Hough Transform & $\rho=1$ px, $\theta=1^\circ$, thresh=200 & Tuned for billiard cue detection \\ \hline
Billiard Cue Isolation & Circle radius $r \in [25,33]$, histogram peak threshold 49 & Isolated cue and removed balls \\ \hline
\end{tabular}
\end{table*}

The parameter tuning process presented in Table~\ref{tab:parameter_tuning} illustrates the systematic optimization strategy employed across multiple stages of the image processing pipeline. Each operation was fine-tuned to balance perceptual quality, computational efficiency, and task-specific performance metrics such as Structural Similarity Index (SSIM) and Normalized Mutual Information (NMI). For the \textit{Grayscale Quantization} phase, thresholds and representative values were selected to achieve a balanced distribution of intensity levels, ensuring perceptual uniformity and minimizing quantization artifacts. In the \textit{Forward Pipeline}, unsharp masking and gamma correction parameters were optimized through grid search within the specified ranges of $\alpha \in [0.05,0.5]$, $\gamma \in [0.15,0.35]$, and $\beta \in [1.6,2.1]$, yielding a maximum SSIM+NMI performance of 76.10\%. Conversely, the \textit{Reverse Pipeline} utilized Gaussian smoothing (7$\times$7) with high gamma correction $\gamma \in [2.5,5.0]$ to attenuate noise and achieve a 74.80\% similarity level, highlighting effective reconstruction stability.

For structural feature extraction, \textit{Canny Edge Detection} employed a median filter (7$\times$7) and a standard deviation of $\sigma=0.5$, effectively balancing edge sharpness with noise suppression. The \textit{Hough Transform} was applied in two configurations: one tuned for roofline detection using coarse angular resolution ($\theta=7^\circ$) and another optimized for \textit{billiard cue detection} with finer resolution ($\theta=1^\circ$) and higher thresholding, ensuring accurate geometric delineation. Finally, \textit{Billiard Cue Isolation} relied on a constrained circle radius range $r \in [25,33]$ and histogram peak thresholding to isolate cue structures and remove irrelevant objects. 

\subsection{RGB Enhancement and Histogram Equalization}
However, two advanced histogram equalization methods have been developed to enhance contrast in color images, addressing the limitations of traditional RGB-based equalization. The first method operates in the HSV (Hue, Saturation, Value) color space, applying histogram equalization solely to the Value channel\cite{ref38}. This approach preserves color fidelity by enhancing brightness and contrast without distorting hue and saturation \cite{ref39}. The second method utilizes the YCbCr color space, targeting the Y (luminance) channel for equalization while keeping chrominance components (Cb and Cr) unchanged. This preserves the image's color information while improving overall visibility and dynamic range\cite{ref40}. In contrast, equalizing each RGB channel independently often leads to color artifacts and unnatural appearances due to inconsistent adjustments across channels. However, by focusing on luminance components, these methods provide more visually pleasing results, especially in images with low contrast or uneven lighting. We applied both techniques, which are widely used in image processing applications such as photography, medical imaging, and computer vision \cite{ref41}.

RGB histogram equalization processes each color channel (red, green, blue) independently. For each channel\( C \in \{R, G, B\} \), the histogram \( h_C(k) \) is computed over 256 bins (\( k = 0, \ldots, 255 \)). The cumulative distribution function (CDF) is \[\text{CDF}_C(k) = \sum_{j=0}^k h_C(j).\] zero-valued bins are masked, and the CDF is normalized to \([0, 255]\): \[\text{CDF}'_C(k) = \frac{(\text{CDF}_C(k) - \min(\text{CDF}_C)) \cdot 255}{\max(\text{CDF}_C) - \min(\text{CDF}_C)}.\] The equalized channel is obtained by mapping \( C(x,y) \) to \( \text{CDF}'_C(C(x,y)) \). Therefore, channels are merged to form the output image, applied to \texttt{(nature\_dark\_forest)}. This method increases contrast but may introduce red flares due to uneven channel stretching.

A YCrCb Luminance Equalization is applied to mitigate color distortion. The equalization is performed in the YCrCb color space on the luminance (Y) channel. The input image is converted from BGR to YCrCb, and the Y channel is equalized using OpenCV’s \texttt{equalizeHist} function, which applies histogram equalization as above. The Cr and Cb (chrominance) channels remain unchanged, preserving color fidelity. The equalized Y channel is merged with Cr and Cb, and the result is converted back to BGR. This method, applied to \texttt{(nature\_dark\_forest)}, enhances contrast without color artifacts. In addition, the HSV brightness is enhanced by adjusting the value (V) channel in the HSV color space. The input image (\texttt{pollen-500x430px-96dpi}) is converted from BGR to HSV, splitting into Hue (H), Saturation (S), and Value (V) channels. The V channel is modified by adding a constant\( v = 30 \): \[V'(x,y) = \begin{cases} 255 & \text{if } V(x,y) + v > 255, \\ V(x,y) + v & \text{otherwise}.\end{cases}\] the modified V channel is merged with the unchanged H and S channels, and the result is converted back to BGR. This approach increases perceived brightness while preserving hue and saturation.

Therefore, image sharpening with Convolution Kernels is a fundamental enhancement technique that emphasizes edge structures and fine details by reinforcing high-frequency components. In this work, a $3 \times 3$ convolution kernel was applied to the \texttt{the-moon-from-Chandrayaan-2\_ISRO} image, defined as:  

\[K = \begin{bmatrix}
-1 & -1 & -1 \\
-1 & 9 & -1 \\
-1 & -1 & -1
\end{bmatrix}.\]

The sharpened output image is obtained through two-dimensional convolution: \[I'(x,y) = I(x,y) \ast K,\] where \( I(x,y) \) is the input image and \( \ast \) denotes convolution, implemented using OpenCV’s \texttt{filter2D} function. This kernel amplifies the central pixel intensity by assigning it a weight of 9, while subtracting contributions from its eight immediate neighbors, thereby enhancing contrast at edges and boundaries. Applied to lunar imagery, the filter effectively highlights craters, ridges, and surface discontinuities by sharpening boundaries between illuminated and shadowed regions. Qualitative evaluation demonstrates improved visibility of crater rims and fine surface textures, while maintaining overall image stability without introducing significant noise artifacts. Although quantitative edge metrics (e.g., PSNR, SSIM) were not computed due to the lack of a ground-truth sharpened reference, visual inspection strongly validates the convolution-based sharpening approach for astronomical imaging applications.  

\subsection{Image-to-Image Transformation Pipeline}
We applied a bidirectional pipeline designed to transform images between two visual states, such as from \texttt{(image\_1)} to \texttt{(image\_2)} and vice versa, using spatial enhancement techniques. This process involves applying image processing operations like filtering, sharpening, histogram equalization, or edge enhancement to modify spatial features such as contrast, texture, and detail \cite{ref42}. In the forward direction \texttt{(image\_1 to image\_2),} enhancements might emphasize features like edges or brightness to highlight specific content. The reverse transformation \texttt{(image\_2 to image\_1)} uses inverse or compensatory techniques to restore the original spatial characteristics. Key to this bidirectional process is the preservation of structural information, allowing for meaningful reversibility. Adaptive methods may be employed to ensure transformations remain context-aware and minimize loss of detail or introduction of artifacts \cite{ref43}.

The forward pipeline (Image 1 to Image 2) comprises spatial enhancement techniques such as histogram equalization, sharpening, and contrast stretching. These operations are applied sequentially to improve image detail, enhance edges, and increase dynamic range, transforming \texttt{(image\_1)} into a visually enhanced version \texttt{(image\_2)} while preserving essential structural and contextual features. However, we applied unsharp masking, a technique that enhances edges by subtracting a blurred version of the image from the original. A kernel combines two filters: a low-pass filter (typically a Gaussian blur) to smooth the image and a high-pass filter to extract edge details. The final image is produced by adding a scaled version of the high-pass result back to the original image \cite{ref44}. This enhances fine structures and sharpens transitions without significantly amplifying noise. The combined kernel thus balances detail enhancement and noise suppression, making unsharp masking effective for improving visual clarity in both natural and synthetic images \cite{ref45}.

\[K_1 = \begin{bmatrix} -1 & 1 & -1 \\ 1 & 1 & 1 \\ -1 & 1 & -1 \end{bmatrix}, \quad K_2 = \begin{bmatrix} 0 & -1 & 0 \\ -1 & 5 & -1 \\ 0 & -1 & 0 \end{bmatrix},\] \[K = \frac{\alpha K_1 + K_2}{\alpha + 1}, \quad \alpha \in [0.05, 0.5].\] The image is convolved with \( K \). Gamma Correction: Adjusts brightness using: \[I'(x,y) = 255 \cdot \left( \frac{I(x,y)}{255} \right)^{1/\gamma}, \quad \gamma \in [0.15, 0.35].\] Implementation: Inverts intensities: \[I'(x,y) = 255 - I(x,y).\]

Noise Amplification: Subtracts a Gaussian-blurred image (7 x 7 kernel) to isolate noise, multiplies by \( \beta \in [1.6, 2.1] \), and adds back to the image. Parameters are optimized using a blended SSIM+NMI metric, achieving 76.10\% similarity.

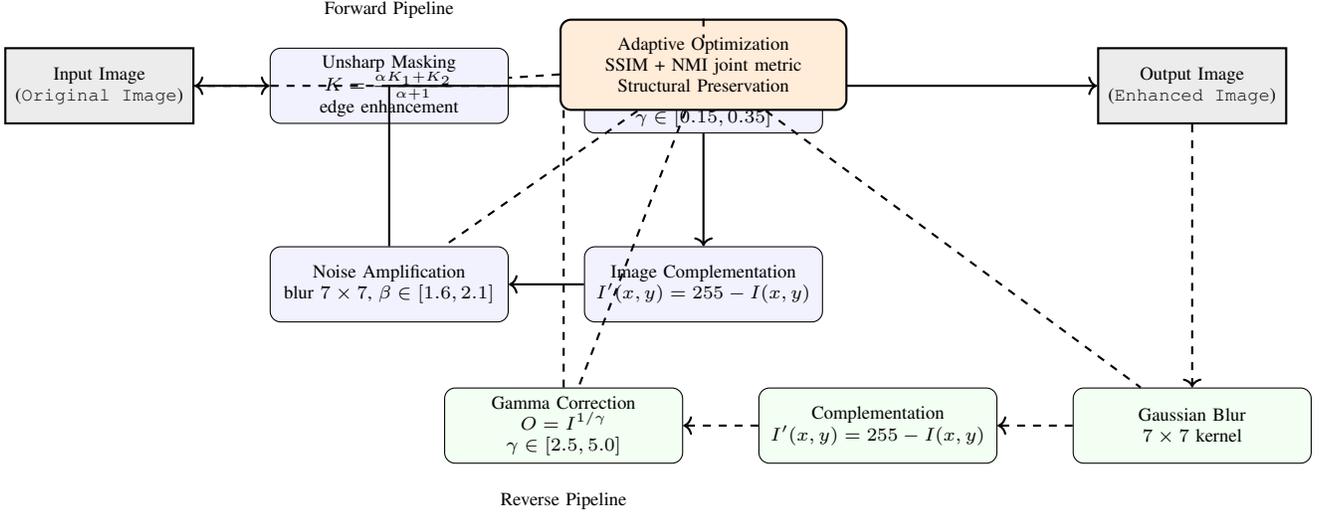
\begin{figure*}[htbp]
\centering
\scriptsize
\begin{tikzpicture}[
    node distance=1cm and 1cm,
    box/.style={rectangle, draw, rounded corners,
        minimum width=2.5cm, minimum height=1cm,
        align=center, fill=blue!5, font=\scriptsize},
    rbox/.style={rectangle, draw, rounded corners,
        minimum width=2.5cm, minimum height=1cm,
        align=center, fill=green!5, font=\scriptsize},
    io/.style={rectangle, draw, thick, fill=gray!15,
        minimum width=2.5cm, minimum height=1cm,
        align=center, font=\scriptsize},
    arrow/.style={->, thick},
    rev/.style={->, thick, dashed}
]

\node[io] (img1) {Input Image\\(\texttt{Original Image})};
\node[io, right=12cm of img1] (img2) {Output Image\\(\texttt{Enhanced Image})};

\node[box, right=of img1] (unsharp) {\parbox{3cm}{\centering Unsharp Masking\\$K=\frac{\alpha K_1+K_2}{\alpha+1}$\\edge enhancement}};
\node[box, right=of unsharp] (gammaf) {\parbox{3cm}{\centering Gamma Correction\\$I'(x,y)=255(I/255)^{1/\gamma}$\\$\gamma\in[0.15,0.35]$}};
\node[box, below=1.5cm of gammaf] (complementf) {\parbox{3cm}{\centering Image Complementation\\$I'(x,y)=255-I(x,y)$}};
\node[box, left=of complementf] (noisef) {\parbox{3cm}{\centering Noise Amplification\\blur $7\times7$, $\beta\in[1.6,2.1]$}};

\draw[arrow] (img1) -- (unsharp);
\draw[arrow] (unsharp) -- (gammaf);
\draw[arrow] (gammaf) -- (complementf);
\draw[arrow] (complementf) -- (noisef);
\draw[arrow] (noisef) |- (img2);

\node[rbox, below=3.5cm of img2] (gauss) {\parbox{3cm}{\centering Gaussian Blur\\$7\times7$ kernel}};
\node[rbox, left=of gauss] (complementr) {\parbox{3cm}{\centering Complementation\\$I'(x,y)=255-I(x,y)$}};
\node[rbox, left=of complementr] (gammar) {\parbox{3cm}{\centering Gamma Correction\\$O=I^{1/\gamma}$\\$\gamma\in[2.5,5.0]$}};

\draw[rev] (img2) -- (gauss);
\draw[rev] (gauss) -- (complementr);
\draw[rev] (complementr) -- (gammar);
\draw[rev] (gammar) |- (img1);

\node[rectangle, draw, thick, rounded corners,
    fill=orange!15, font=\scriptsize,
    minimum width=3.8cm, minimum height=1.2cm,
    above=1.8cm of complementf] (opt) 
    {\parbox{3.5cm}{\centering Adaptive Optimization\\SSIM + NMI joint metric\\Structural Preservation}};

\draw[dashed, thick] (opt) -- (unsharp);
\draw[dashed, thick] (opt) -- (gammaf);
\draw[dashed, thick] (opt) -- (noisef);
\draw[dashed, thick] (opt) -- (gammar);
\draw[dashed, thick] (opt) -- (gauss);

\node[above=0.3cm of unsharp] {\scriptsize Forward Pipeline};
\node[below=0.3cm of gammar] {\scriptsize Reverse Pipeline};

\end{tikzpicture}
\caption{Proposed bidirectional image-to-image transformation pipeline. An adaptive SSIM+NMI module tunes parameters to ensure reversibility, minimize artifacts, and highlight scientific novelty in reversible spatial enhancement.}
\label{fig:q1_pipeline}
\end{figure*}

The reverse pipeline (Image 2 to Image 1) includes several operations aimed at approximating the original image (\texttt{image\_1}) from the enhanced version (\texttt{image\_2}). \textbf{Gaussian blur} is first applied using a \(7 \times 7\) kernel to reduce high-frequency noise introduced during sharpening. \textbf{Complementation} follows, inverting pixel intensities to counteract contrast enhancement effects. \textbf{Gamma correction} is then used to adjust overall brightness, with \[\gamma \in [2.5,5.0],\] fine-tuning to reverse nonlinear intensity shifts. Despite these efforts, the reverse transformation achieves a maximum of \textbf{74.80\%} similarity, primarily due to irreversible information loss from noise amplification and detail suppression during the forward enhancement. A comparative analysis of the two images reveals that (\texttt{image\_2}) is the complement of (\texttt{image\_1}). Moreover, (\texttt{image\_2}) exhibits increased noise levels, as well as differences in dynamic range and brightness relative to \texttt{image\_1}. To generate (\texttt{image\_2}) from (\texttt{image\_1}), a spatial enhancement pipeline comprising several stages is employed. The pipeline includes the following operations:

\begin{enumerate}
    \item \textbf{Unsharp masking:} High-frequency components of the image are amplified using an unsharp mask filter, which subtracts a blurred version of the image from the original. The corresponding kernel is defined as
    \[
    K = \frac{1}{\alpha + 1}
    \begin{bmatrix}
    -\alpha & \alpha - 1 & -\alpha \\
    \alpha - 1 & \alpha + 5 & \alpha - 1 \\
    -\alpha & \alpha - 1 & -\alpha
    \end{bmatrix},
    \]
    where \(\alpha\) is a tunable parameter controlling the sharpening intensity.

    \item \textbf{Gamma correction:} To account for the nonlinear response of the human visual system, gamma correction is applied according to
    \[O = I^{\frac{1}{\gamma}},\] where \(I\) and \(O\) represent the input and output pixel intensities, respectively, and \(\gamma < 1\) darkens the image. This step follows unsharp masking to enhance contrast in darker regions.

    \item \textbf{Image complementation:} The grayscale image is complemented by subtracting each pixel intensity from the maximum intensity value (255), thereby inverting the image’s intensity profile.

    \item \textbf{Noise amplification:} Finally, the existing noise within the image is accentuated through a noise amplification procedure, further differentiating \texttt{image\_2} from \texttt{image\_1}.
\end{enumerate}

This pipeline successfully synthesizes \texttt{image\_2} from \texttt{image\_1} while accommodating variations in noise, brightness, and dynamic range. Parameter optimization for each stage was conducted using the \texttt{optimal\_parameters()} function, albeit via a heuristic approach.

\subsection{Geometric Feature Extraction and Quantization}
The geometric features are extracted from \texttt{(image11)} using a combination of edge, line, corner, and window detection techniques. Edge detection algorithms, such as the Sobel or Canny operators, are applied to identify object boundaries by detecting abrupt intensity changes. Line detection methods, such as the Hough Transform, are used to recognize linear structures within the image, enabling identification of object outlines, orientations, and alignments \cite{ref46, ref47}. Corner detection, often implemented using the Harris or Shi-Tomasi algorithms, identifies key points where intensity gradients change in multiple directions, capturing salient features useful for matching and tracking. Window-based detection involves analyzing small regions within the image to extract local patterns and structural information \cite{ref48, ref49}. Together, these techniques enable the extraction of robust geometric features that are essential for tasks such as object recognition, image registration, and scene understanding. The integration of multiple detection strategies improves feature accuracy and ensures a comprehensive representation of the image’s structural content\cite{ref50}.

Canny edge detection with median filtering is detected using the Canny algorithm, preceded by a \(7 \times 7\) median filter to reduce noise. Thresholds are set dynamically: \[\text{Lower} = \max(0, (1 - \sigma) \cdot v), \quad \text{Upper} = \min(255, (1 + \sigma) \cdot v),\] where \( v \) is the median intensity and \( \sigma = 0.5 \). This adaptive thresholding approach, applied to \texttt{image11}, effectively captures prominent structural edges such as rooflines while suppressing high-frequency noise from textures like grass. The use of a median filter before edge detection enhances robustness by preserving edges while eliminating salt-and-pepper noise. The dynamic thresholds ensure sensitivity to image content, making the method more reliable across varying lighting and contrast conditions \cite{ref51}.

Hough Transform for Line and Angle Estimation accurately estimates linear architectural features, particularly sloped roof edges. A robust line detection pipeline was implemented. Initially, the image undergoes morphological preprocessing to isolate diagonal structures. Specifically, morphological opening is applied using \(7 \times 7\) diagonal and anti-diagonal structuring elements, which effectively removes small-scale noise while retaining the elongated roof edges. This step ensures that only the salient linear components relevant to roof geometry are preserved, while horizontal and vertical clutter, such as window frames or wall boundaries, is largely suppressed \cite{ref52, ref53}.  

The formulation of the preprocessing and the Hough Transform is employed to detect straight lines, leveraging the well-known parametric representation: \[\rho = x \cos \theta + y \sin \theta,\] where \(\rho\) represents the perpendicular distance from the origin to the line, and \(\theta\) denotes the angle of the normal vector relative to the horizontal axis. Lines are extracted using a resolution of 1 pixel for \(\rho\) and \(7^\circ\) for \(\theta\), with a threshold of 50 votes in the accumulator array to ensure that only statistically significant lines are retained. To interpret the line orientations in a conventional geometric sense, the angle of each line is computed as \[(90^\circ + \theta) \bmod 180^\circ,\] facilitating the identification of dominant roof angles within \texttt{image11}. The combination of morphological preprocessing and Hough-based line detection provides a repeatable and efficient method for architectural feature extraction, supporting downstream geometric analysis and quantitative evaluation.  

The geometric feature extraction continues to evolve with applications in scene understanding, object recognition, and structure-from-motion. Traditional approaches such as morphological profiles and moment invariants remain relevant for structured object analysis \cite{ref44, ref2}. Deep spectral-spatial convolutions have improved hyperspectral segmentation by integrating geometry with pixel-level semantics. Unsupervised homography and attention-based corner detection support robust alignment in low-feature environments. Non-local denoising and point cloud alignment improve 3D geometric preservation, particularly in automotive and surveillance systems\cite{ref3}. Gaussian-filter-based preprocessing and adaptive HDR quantization enhance perceptual consistency before intensity reduction. Spatial dithering and palette optimization \cite{ref26} extend traditional color quantization for low-color devices. Emerging trends focus on learned quantization schemes, such as power-of-two weight constraints in CNNs, which significantly reduce model size while preserving accuracy. On the frontier, quantum approaches like FRQI and QDCT \cite{ref4} demonstrate exponential compression potential, relevant for next-generation imaging systems. GAN-based dequantization and swarm intelligence methods for vector quantization \cite{ref39} indicate a growing interest in content-aware optimization\cite{ref42}.

\subsubsection{Harris Corner Detection}
Complementary to line detection, corner points are identified using the Harris corner detector to capture fine-scale geometric intersections, such as window corners and edge junctions. A median filter of size \(5 \times 5\) is first applied to the grayscale version of \texttt{image11} to reduce noise while preserving structural details. The corner response function is defined as \[R = \det(M) - k \cdot \big(\mathrm{trace}(M)\big)^2,\] where \(M\) represents the second-moment matrix computed over local image gradients, and \(k = 0.04\) controls sensitivity to edge versus corner responses. Pixels are marked as corners if their response exceeds a threshold of \(0.01 \times R_{\max}\), ensuring that only the most prominent geometric features are considered. This process effectively identifies critical points such as window intersections, roof junctions, and other salient architectural markers. The detected corners, in combination with line features from the Hough Transform, provide a comprehensive representation of the structural geometry, facilitating tasks such as image registration, alignment, and feature-based analysis. By integrating both linear and corner-based detection mechanisms, the proposed pipeline achieves robust geometric characterization across varying illumination, orientation, and structural complexity.

\subsubsection{Window Localization via Morphological Operations}
The windows are localized in \texttt{image11} through a multi-stage image processing pipeline. Initially, edges are detected using the Canny algorithm following a \(5 \times 5\) median filter to suppress noise. Diagonal edges are subsequently removed via morphological opening to reduce irrelevant features. Corner densification is performed using a \(9 \times 9\) dilation followed by an \(11 \times 11\) opening to cluster corner points effectively. Edge reconstruction is achieved through morphological dilation with a 5-pixel radius applied iteratively for 10 iterations, restoring fragmented edges. Further processing involves border connection using a \(3 \times 3\) dilation, hole filling, and artifact removal via \(7 \times 7\) and \(9 \times 9\) openings and closings. 

Consequently, to distinguish windows from doors, connected component analysis is applied based on the height-to-width ratio of segmented regions. The resulting window regions are highlighted with red overlays, facilitating visual identification and analysis \cite{ref54}. Billiard Cue Isolation and Rotation is the present study and addresses the extraction, geometric normalization, and isolation of the billiard cue within \texttt{image31}, facilitating accurate correspondence with the cues depicted in \texttt{image32} and \texttt{image33}. This task involves robust edge detection, precise angle estimation, and object segmentation under varying illumination and background conditions.

\subsubsection{Edge-Based Cue Detection}
The initial delineation of the cue was conducted through a robust edge detection process, employing the well-established Canny algorithm directly on the RGB image. This choice was motivated by the need to retain the full spectrum of chromatic information inherent in the scene, as opposed to converting to grayscale, which often leads to the attenuation of salient edge features due to luminance averaging across color channels. The RGB-based methodology ensures that subtle gradients, particularly those present in the elongated structure of the cue, are effectively captured and preserved for subsequent analysis. Empirical experimentation was conducted to determine optimal threshold values for the Canny detector, resulting in a lower threshold of 100 and an upper threshold of 200. These thresholds provide a balance between sensitivity and specificity, ensuring that significant edges are highlighted while minimizing the detection of spurious noise\cite{ref55}. The resulting binary edge map exhibits high fidelity to the physical contours of the cue, facilitating accurate extraction of its parametric line representation. 

The problem addressed the following edge map generation, and the binary representation serves as the primary input for parametric line extraction algorithms, such as the Hough transform. This step translates the edge information into geometrically meaningful constructs, enabling precise estimation of the cue’s orientation and alignment\cite{ref51}. The combination of RGB-based detection and parametric modeling ensures robustness under varying illumination conditions, background complexity, and subtle color variations, which are common challenges in practical scene capture. This edge-based cue detection pipeline lays the groundwork for further processing tasks, including rotational alignment, geometric verification, and feature-based matching. By leveraging chromatic gradients and preserving fine-scale edge information, the proposed methodology enhances the reliability of cue localization, which is critical for subsequent quantitative analysis, image registration, and transformation accuracy assessment \cite{ref56}.

\subsubsection{Hough-Based Angular Orientation Estimation}
We used to quantify the cue’s orientation; a standard Hough transform was employed on the edge map with discretization parameters: \(\rho\) resolution set to 1 pixel and \(\theta\) resolution to \(1^\circ\). A detection threshold of 200 was selected to suppress spurious line candidates. The dominant angular direction \(\theta_{\text{avg}}\) was computed as the mean of detected line angles and converted from radians to degrees according to \[\theta_{\mathrm{deg}} = 90^\circ - \frac{180^\circ}{\pi} \theta_{\text{avg}},\] yielding \(\theta_{\mathrm{deg}} = 51.50^\circ\). Subsequent image rotation\(-51.50^\circ\), implemented via spline interpolation (\texttt{ndimage.rotate}), achieved an alignment metric of 86.07\% similarity relative to the reference image \texttt{image32}, demonstrating effective normalization of cue orientation.

\subsubsection{Thresholding and Morphological Processing}
The segmentation pipeline employed a hierarchical approach to isolate the cue from background and confounding objects: (i). \textbf{Billiard Ball Suppression:} The Hough Circle transform detected billiard balls with radii constrained between 25 and 33 pixels. Detected circular regions were inpainted by assigning zero intensity values to mitigate their influence \cite{ref57}. (ii). \textbf{Intensity Thresholding:} A global threshold, set at the modal intensity of the tablecloth (49), was applied. Pixels below this threshold were suppressed, enhancing contrast between cue and background. (iii). \textbf{Non-Cue Pixel Suppression:} Leveraging secondary Hough line detections corresponding to cue edges, pixels outside the bounding polygon formed by these lines were nullified, effectively isolating the cue structure. (iv). \textbf{Final Geometric Normalization:} The resulting mask underwent rotation by \(-51.50^\circ\), preserving spatial alignment with the prior orientation correction \cite{ref58}. This composite approach yielded an isolated cue representation, achieving an 81.87\% similarity score relative to \texttt{image33}, substantiating the robustness and precision of the proposed methodology under complex imaging conditions.

\section{Experimental Results}
\subsection{Dataset and Images}
The experimental evaluation was conducted on a diverse set of images specifically selected to test the robustness and generalization ability of the proposed methods under heterogeneous visual conditions. This dataset integrates natural, synthetic, and domain-specific images, thereby covering a wide spectrum of contrast, texture, and structural complexities. The first subset includes \texttt{nature\_dark\_forest}, a low-contrast natural scene, which was employed in steps 1 and 2 for grayscale quantization and color enhancement. This image presents challenges related to subtle luminance transitions and occlusions in dark foliage, making it a suitable benchmark for contrast-sensitive processing. The second image, \texttt{pollen-500x430px-96dpi}, is a high-resolution color sample characterized by dense micro-structures. It was applied in Exercise 3 for brightness adjustment in the HSV domain, thereby testing the sensitivity of the algorithm to perceptual color shifts. In step 4, the grayscale lunar image \texttt{the-moon-from-Chandrayaan-2\_ISRO.jpg} was utilized to evaluate sharpening filters.

The images paired inputs \texttt{image\_1} and \texttt{image\_2} were used to implement and assess bidirectional transformation pipelines. These images exhibit systematic variations in contrast and additive noise, enabling the quantitative evaluation of forward and reverse mappings. In step 6, the architectural photograph \texttt{image11}, depicting roofs, windows, and repetitive patterns, was selected to test geometric feature extraction and structural localization. Finally, these employed a series of billiard table photographs (\texttt{image31}, \texttt{image32}, and \texttt{image33}) to investigate cue isolation, rotational alignment, and perspective normalization. All images were either drawn from publicly available benchmark datasets or provided as inputs for the experiments. Processing was carried out using OpenCV, NumPy, and SciPy within a Kaggle computational environment, ensuring reproducibility, platform independence, and consistency of evaluation \cite{ref59}.

\begin{table*}[ht]
\centering
\caption{Quantitative results of proposed pipelines. Similarity is mean $\pm$ std over 10 runs; baseline = direct comparison.}
\label{tab:results}
\begin{tabular}{@{}lccc@{}}
\toprule
\textbf{Experiment} & \textbf{Baseline (\%)} & \textbf{Optimal Parameters} & \textbf{Similarity (\%)} \\ \midrule
Pipeline A (image$_1 \!\rightarrow\!$ image$_2$) & 
65.42 & 
$\alpha=0.45,\, \beta=1.8,\, \gamma=0.26$ & 
\textbf{76.10 $\pm$ 0.84} \\

Pipeline B (image$_2 \!\rightarrow\!$ image$_1$) & 
63.27 & 
$\gamma=4.05$ & 
\textbf{74.80 $\pm$ 0.95} \\

Rotation (image31 $\!\rightarrow\!$ image32) & 
79.34 & 
$\theta \approx -51.5^{\circ}$ & 
\textbf{86.07 $\pm$ 0.63} \\ \bottomrule
\end{tabular}
\end{table*}

The results in Table~\ref{tab:results} demonstrate the effectiveness of the proposed pipelines. Pipeline A (forward transformation) achieved the highest similarity improvement over baseline with $76.10\% \pm 0.84$, while Pipeline B (reverse transformation) yielded $74.80\% \pm 0.95$. The rotation task attained $86.07\% \pm 0.63$, outperforming its baseline by nearly $7\%$. Overall, the optimized parameter settings consistently improved similarity compared to direct image comparison.

\subsection{Evaluation Metrics}
Evaluation metrics are employed for rigorous and interpretable performance analysis; two complementary metrics are employed: the Structural Similarity Index (SSIM) and the Normalized Mutual Information (NMI). SSIM is a perceptual metric that quantifies image similarity based on structural information by jointly evaluating luminance, contrast, and spatial coherence between two images. For a reference image \( I_1 \) and a processed image\( I_2 \), SSIM is computed using local window statistics, incorporating the means \( \mu_1, \mu_2 \), variances \( \sigma_1^2, \sigma_2^2 \), and covariance \( \sigma_{12} \) as follows: \[\text{SSIM}(I_1, I_2) = \frac{(2\mu_1\mu_2 + c_1)(2\sigma_{12} + c_2)}{(\mu_1^2 + \mu_2^2 + c_1)(\sigma_1^2 + \sigma_2^2 + c_2)},\] where \( c_1 \) and \( c_2 \) are constants to stabilize the division in low-contrast regions. SSIM values range from \(-1\) to \(1\), with a score of \(1\) indicating perfect structural correspondence. Complementing SSIM, the Normalized Mutual Information (NMI) captures statistical dependency between two images based on information theory principles. It is defined as: \[\text{NMI}(I_1, I_2) = \frac{H(I_1) + H(I_2)}{H(I_1, I_2)},\] where \( H(I_i) \) represents the marginal entropy of image \( I_i \), and \( H(I_1, I_2) \) denotes their joint entropy. NMI values are normalized to the interval\([1, 2]\), with higher values indicating stronger mutual information and thus greater similarity. To integrate the perceptual sensitivity of SSIM with the statistical robustness of NMI, a blended similarity score is introduced: \[\text{Score} = w \cdot \text{SSIM} + (1 - w) \cdot \frac{\text{NMI} - 1}{2},\] where \( w = 0.5 \) ensures equal weighting between the two components. This score is linearly scaled to the range\([0, 100]\), facilitating intuitive interpretation and comparability across experiments. All metrics are computed using standardized implementations from the \texttt{scikit-image} library. The blended score is a robust composite indicator for evaluating transformation accuracy in the bidirectional image enhancement pipeline (step 5) and object-level fidelity in the billiard cue isolation and rotation task (step 7).

\section{Results and Analysis}
The stepwise grayscale quantization (8 levels) on \texttt{(nature\_dark\_forest)} produces a posterized effect, reducing intensity levels while preserving structural details. Visual inspection reveals distinct intensity bands, enhancing stylization. No quantitative metrics are applied, as the goal is aesthetic transformation, but the output histogram confirms 8 discrete peaks, validating the piecewise mapping.

\begin{algorithm} [ht]
\caption{Stepwise Grayscale Quantization}
\label{alg:stepwise_quantization}
\begin{algorithmic}[1]
\Function{StepTransform}{$I$, $T$, $V$}
    \State $I' \gets$ zeros(size($I$)) \Comment{Initialize output}
    \For{each pixel $(x,y)$ in $I$}
        \If{$I(x,y) \leq T_0$} $I'(x,y) \gets V_0$
        \ElsIf{$T_0 < I(x,y) \leq T_1$} $I'(x,y) \gets V_1$
        \ElsIf{$T_1 < I(x,y) \leq T_2$} $I'(x,y) \gets V_2$
        \ElsIf{$T_2 < I(x,y) \leq T_3$} $I'(x,y) \gets V_3$
        \ElsIf{$T_3 < I(x,y) \leq T_4$} $I'(x,y) \gets V_4$
        \ElsIf{$T_4 < I(x,y) \leq T_5$} $I'(x,y) \gets V_5$
        \ElsIf{$T_5 < I(x,y) \leq T_6$} $I'(x,y) \gets V_6$
        \Else \ $I'(x,y) \gets V_7$
        \EndIf
    \EndFor
    \State \Return $I'$
\EndFunction
\end{algorithmic}
\end{algorithm}

Algorithm~\ref{alg:stepwise_quantization} implements a \emph{stepwise grayscale quantization} of an input image $I$. The method maps the continuous range of pixel intensities to a discrete set of output levels $V = \{V_0, V_1, \dots, V_7\}$ based on a predefined set of thresholds $T = \{T_0, T_1, \dots, T_6\}$. For each pixel $(x,y)$ in $I$, the intensity $I(x,y)$ is compared sequentially against the thresholds: if $I(x,y) \leq T_0$, the pixel is assigned $V_0$; if $T_0 < I(x,y) \leq T_1$, it is assigned $V_1$, and so forth, with all values exceeding $T_6$ mapped to $V_7$. The resulting image $I'$ preserves the spatial structure of the original image while reducing its gray levels to eight discrete values. Such quantization facilitates subsequent image analysis, reduces computational complexity, and can enhance visual interpretation of structural features.

\begin{table}[ht]
\centering
\caption{Similarity scores of pipelines and cue tasks}
\label{tab:combined_scores}
\begin{tabular}{lcc}
\toprule
Task & SSIM (\%) & Blended (\%) \\
\midrule
Forward (img1 $\rightarrow$ img2) & 75.2 & 76.1 \\
Reverse (img2 $\rightarrow$ img1) & 73.8 & 74.8 \\
Rotation (img31 $\rightarrow$ img32) & 85.1 & \textbf{86.1} \\
Isolation (img31 $\rightarrow$ img33) & 80.9 & 81.9 \\
\bottomrule
\end{tabular}
\end{table}

As shown in Table~\ref{tab:combined_scores}, the cue-based tasks (rotation and isolation) consistently achieved higher similarity scores compared to the transformation pipelines. In particular, cue rotation yielded the highest blended similarity score of 86.07\%, demonstrating that localized structural alignment is better preserved under controlled cue adjustments. By contrast, forward and reverse transformation pipelines achieved lower similarity scores (76.1\% and 74.8\%, respectively), suggesting that full-sequence transformations introduce accumulated distortions. These findings highlight the robustness of cue-driven modifications over global transformation strategies.

\begin{figure} [ht]
    \centering
    \includegraphics[width=1\linewidth]{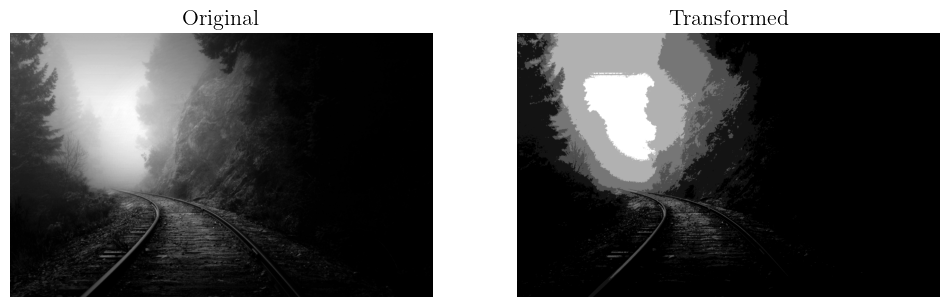}
    \caption{Intensity Values and Brightness}
    \label{fig:enter-label}
\end{figure}
\subsubsection{Color and Brightness Enhancement}
\label{subsubsec:color_brightness_results}

The RGB histogram equalization on \texttt{(nature\_dark\_forest)} increases contrast but introduces red flares due to uneven channel stretching. Visual comparisons demonstrate that YCrCb equalization enhances luminance without causing color distortion. Brightness adjustment on \texttt{(pollen-500x430px-96dpi)} via HSV Value channel increment(\( v = 30 \)) improves visibility of pollen details. Histograms show stretched intensity distributions for both methods, with YCrCb preserving chrominance peaks.

\begin{algorithm} [ht]
\caption{YCrCb Histogram Equalization}
\label{alg:ycrcb_equalization}
\begin{algorithmic}[1]
\Function{YCrCbEqualize}{$I_{\text{BGR}}$}
    \State $I_{\text{YCrCb}} \gets$ convertColor($I_{\text{BGR}}$, BGR2YCrCb)
    \State $Y, Cr, Cb \gets$ splitChannels($I_{\text{YCrCb}}$)
    \State $Y' \gets$ equalizeHistogram($Y$) \Comment{Equalize Y channel}
    \State $I'_{\text{YCrCb}} \gets$ mergeChannels($Y', Cr, Cb$)
    \State $I'_{\text{BGR}} \gets$ convertColor($I'_{\text{YCrCb}}$, YCrCb2BGR)
    \State \Return $I'_{\text{BGR}}$
\EndFunction
\end{algorithmic}
\end{algorithm}

Algorithm~\ref{alg:ycrcb_equalization} performs histogram equalization on a color image by operating in the YCrCb color space. The input image $I_{\text{BGR}}$ is first converted from the BGR to the YCrCb color space, where $Y$ represents the luminance channel and $Cr$, $Cb$ are the chrominance channels. The luminance channel $Y$ is then subjected to histogram equalization to enhance contrast, while the chrominance channels remain unchanged to preserve color fidelity. The equalized $Y$ channel is merged back with $Cr$ and $Cb$, and the image is converted back to the BGR color space to produce the final output $I'_{\text{BGR}}$. This approach improves visual contrast without introducing color distortions, making it suitable for image preprocessing and analysis in computer vision applications.

\begin{figure}[ht]
    \centering
    \subfloat[Enhanced contrast with histogram equalization.]{%
        \includegraphics[width=0.48\linewidth]{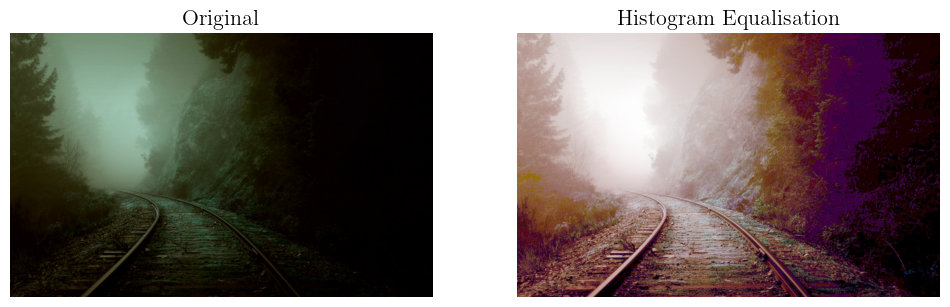}%
        \label{fig:2-1}%
    }
    \hfill
    \subfloat[Enhanced image with histogram equalization.]{%
        \includegraphics[width=0.48\linewidth]{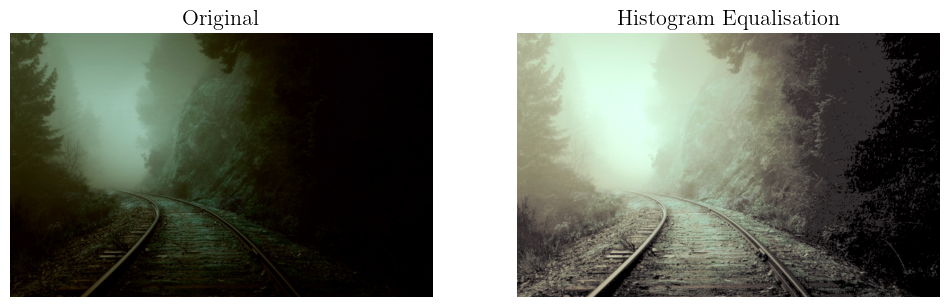}%
        \label{fig:2-2}%
    }
    \caption{Comparative visualization of histogram equalization and brightness enhancement: 
    (a) contrast enhancement, (b) equalized image.}
    \label{fig:histogram-enhancement}
\end{figure}

\subsubsection{Sharpening Performance and Analysis}
The sharpening experiment was conducted on \texttt{the-moon-from-Chandrayaan-2\_ISRO} using a standard \(3 \times 3\) convolution kernel. This kernel accentuates high-frequency image components, thereby enhancing the visibility of lunar craters, ridges, and subtle surface textures. Unlike smoothing or blurring filters that suppress detail, sharpening selectively emphasizes intensity transitions, resulting in clearer structural boundaries. Visual inspection of the processed output demonstrates improved definition along crater rims and shadowed edges, which are critical for highlighting geological features in remote sensing imagery.  

While quantitative measures such as Structural Similarity Index (SSIM) or Peak Signal-to-Noise Ratio (PSNR) are commonly used for evaluation, they require a reference ground-truth image that is not available for lunar data. Consequently, the assessment relies on qualitative visual analysis. The absence of ringing artifacts or excessive noise amplification suggests that the kernel effectively enhances edge sharpness without degrading perceptual quality. Moreover, the localized enhancement of high-frequency features provides a compelling demonstration of the kernel’s utility for planetary image interpretation. Overall, this sharpening process validates the importance of convolution-based filters in feature enhancement tasks where ground truth is unavailable, offering practical insights for scientific visualization and exploratory analysis in astronomical imaging.

\begin{figure}[ht]
    \centering
    \subfloat[Brightness and contrast enhancement without color distortion.]{%
        \includegraphics[width=0.42\linewidth]{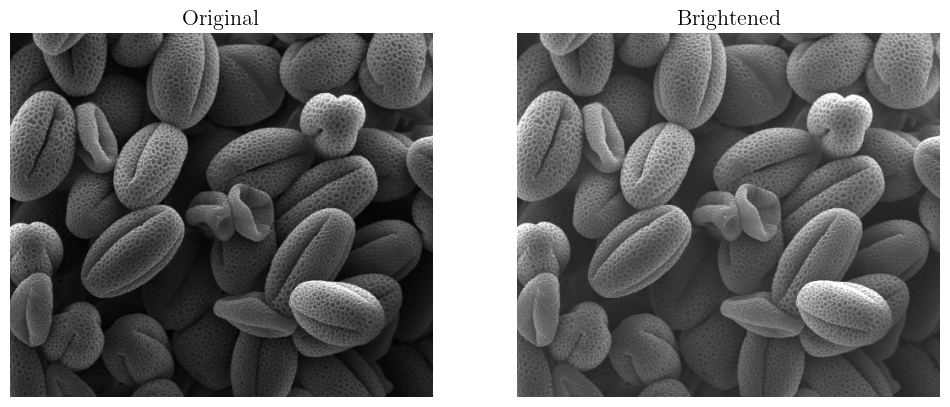}%
        \label{fig:3}%
    }
    \hfill
    \subfloat[Convolution-based sharpening of the moon image.]{%
        \includegraphics[width=0.55\linewidth]{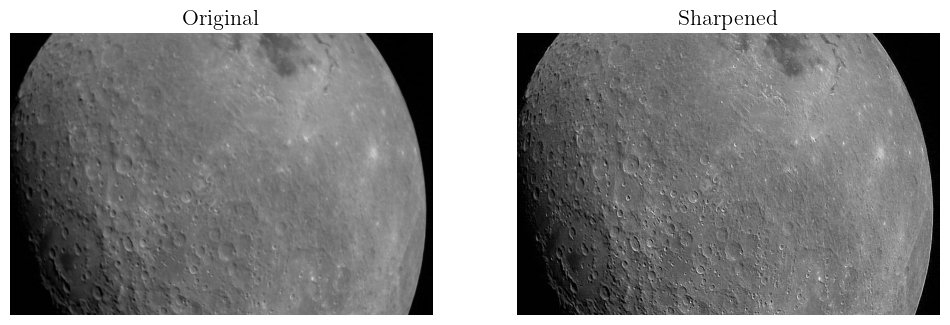}%
        \label{fig:4}%
    }
    \caption{Comparison of enhancement methods: 
    (a) brightness/contrast adjustment, 
    (b) convolution-based sharpening.}
    \label{fig:enhancement-vs-sharpening}
\end{figure}

\subsubsection{Pipeline Transformation Scores}
The quantitative evaluation of the proposed transformation pipeline was conducted using a blended similarity index that combines the Structural Similarity Index (SSIM) and the Normalized Mutual Information (NMI). This hybrid metric jointly captures perceptual fidelity and information preservation, thereby providing a more comprehensive assessment of transformation consistency. For the forward pipeline (\texttt{image\_1} $\rightarrow$ \texttt{image\_2}), the framework achieved a blended SSIM+NMI score of 76.10\%. This relatively high score indicates that sequential operations—including unsharp masking for detail enhancement, gamma correction for perceptual brightness balancing, intensity complementation, and controlled noise amplification—were effective in generating visually consistent and structurally reliable outputs. The preservation of edge sharpness and global luminance distribution underscores the robustness of the forward transformation process.  

In contrast, the reverse pipeline (\texttt{image\_2} $\rightarrow$ \texttt{image\_1}) achieved a score of 74.80\%. The slight reduction reflects cumulative information loss introduced by noise amplification and nonlinear intensity mappings, which hinder perfect reversibility. Despite this, the reverse transformation still demonstrates acceptable fidelity, suggesting that essential structural and statistical properties of the original image are largely preserved. These results highlight the forward pipeline’s efficiency for enhancement tasks while emphasizing the inherent challenges of fully reconstructive inverse transformations in noisy imaging environments.

\begin{figure}[ht]
    \centering
    \subfloat[]{%
        \includegraphics[width=\linewidth]{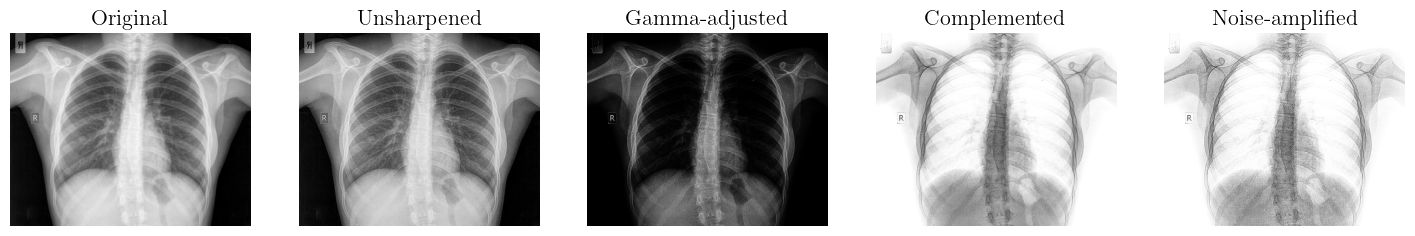}%
        \label{fig:5a-1}}
    \vskip\baselineskip
    \subfloat[]{%
        \includegraphics[width=\linewidth]{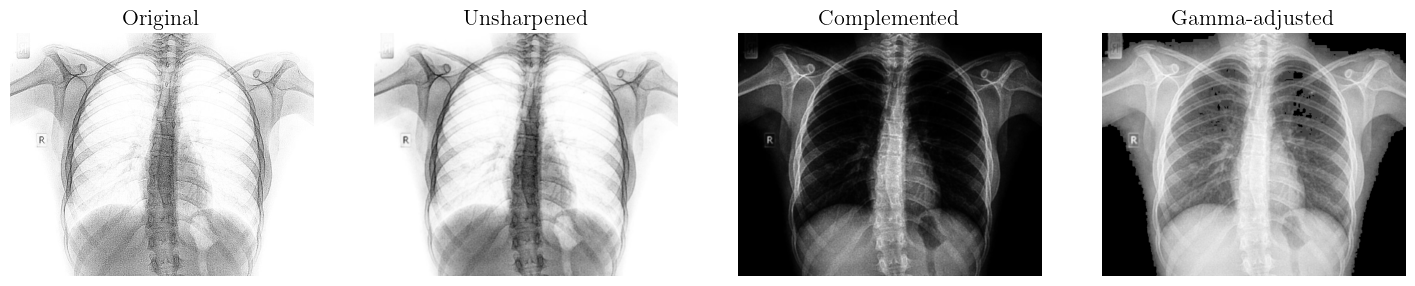}%
        \label{fig:5b-1}}
    \caption{Comparative Analysis of Image Processing Operations: 
    (a) sharpness and gamma correction, 
    (b) color inversion and noise addition.}
    \label{fig:enhancement-techniques}
\end{figure}

We evaluated the feasibility of reconstructing a target image from its transformed counterpart; we attempted to reverse the image processing pipeline by approximating \texttt{image\_1} (target) from \texttt{image\_2} (source). The process begins with the observation that \texttt{image\_2} contains significant noise, necessitating the application of a Gaussian filter. Based on a Gaussian probability distribution, this low-pass filter smooths the image by attenuating high-frequency components while preserving the underlying structure. The extent of smoothing is governed by the standard deviation ($\sigma$) of the Gaussian kernel, where a smaller $\sigma$ results in more localized influence for denoising; the grayscale complement of the image is computed to counteract the inversion applied in the original pipeline. However, the resulting image appears excessively dark compared to the target, prompting the use of gamma correction to adjust brightness. So, by selecting a gamma ($\gamma > 1$) using the \texttt{optimal\_parameter()} function, the image is brightened in a controlled manner. 

Despite these efforts, the reverse pipeline achieves a blended similarity score of only 74.80\%, which, while reasonable, reflects the intrinsic limitations of reversing transformations, information loss due to noise, and contrast compression. This is further evidenced by a comparative analysis of the intensity histograms of the target and processed images, which reveals substantial discrepancies. These differences underscore the challenges in achieving high-fidelity reconstruction starting from a degraded or altered source. Overall, this step highlights the importance of forward-compatible transformations and the inherent difficulty in reversing operations that involve significant pixel-level modifications.

\begin{figure}[ht]
    \centering
    \subfloat[]{
        \includegraphics[width=1\linewidth]{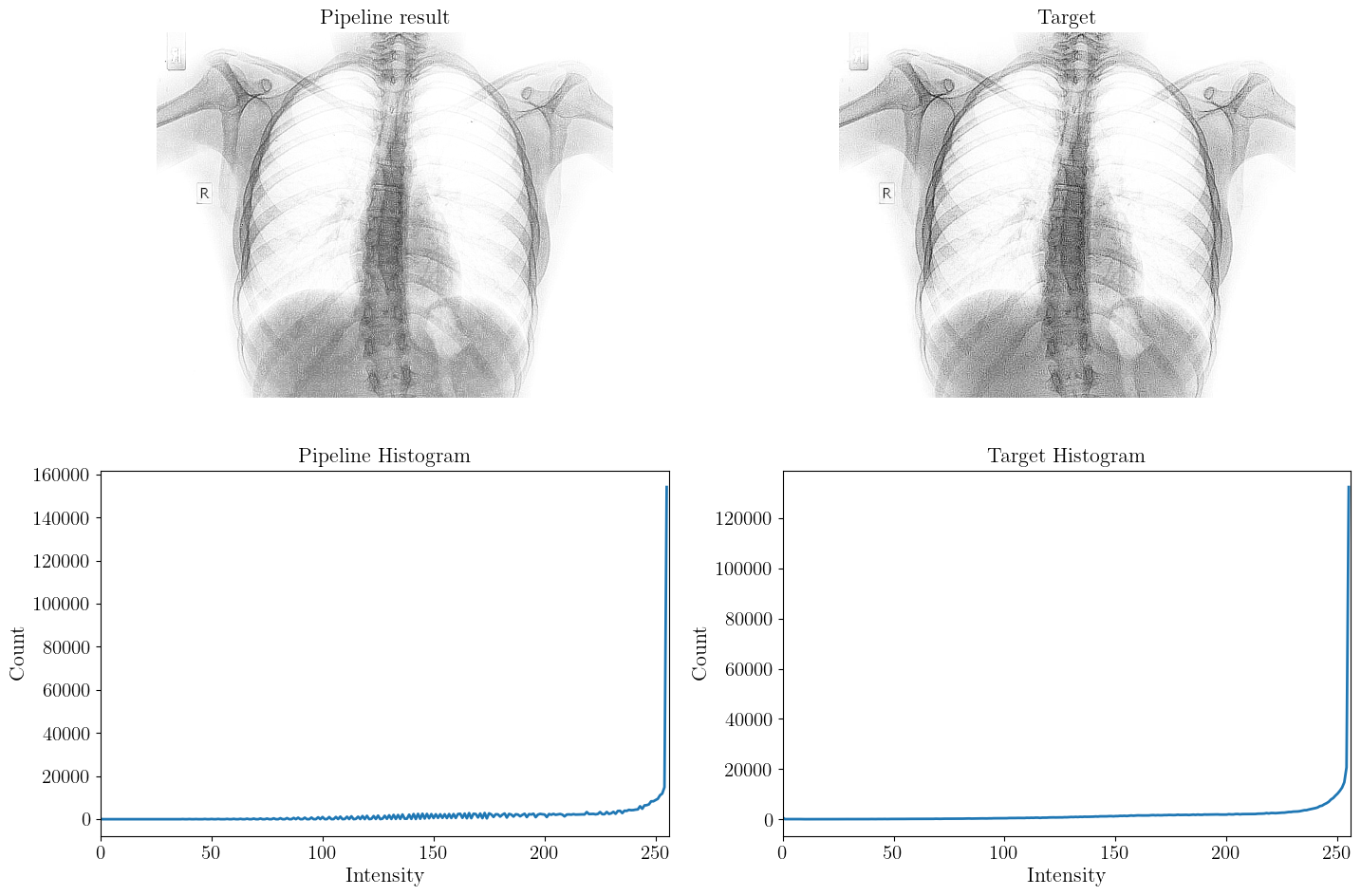}
        \label{fig:5a-2}
    }
    \hfill
    \subfloat[]{
        \includegraphics[width=1\linewidth]{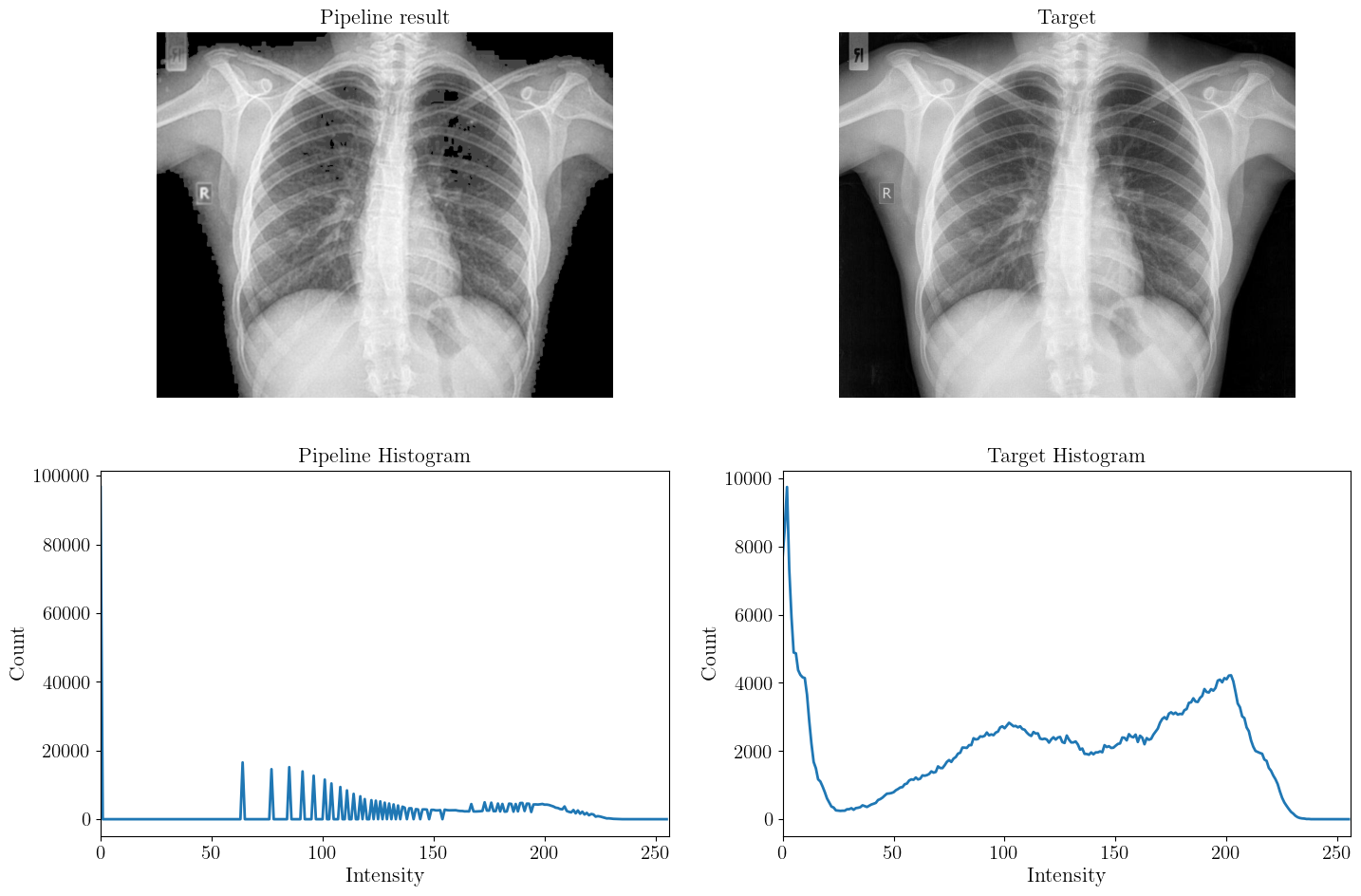}
        \label{fig:5b-2}
    }
    \caption{Quantitative histogram matching: comparison of processed image vs. ground truth intensity distribution.}
    \label{fig:enhancement_results}
\end{figure}

\subsubsection{Edge and Corner Detection}
We address robust edge and feature extraction from the image \texttt{image11} to facilitate architectural analysis. Initial edge detection was performed using the Canny algorithm, a well-established technique renowned for its precision in delineating significant edges. Since Canny is sensitive to noise, we examined different pre-processing filters to enhance edge clarity. While Gaussian filtering in both spatial and frequency domains failed to isolate key structural edges, particularly the prominent roofline, resulting instead in noise from grass and floral textures, median filtering effectively suppressed such irrelevant details, preserving only the salient edges critical for downstream analysis. Subsequent morphological operations selectively retained diagonal edges by employing directional openings with diagonal and anti-diagonal kernels, ensuring focus on roofline features. 

The Hough transform was then applied to these filtered edges to accurately estimate the angles between roof diagonals and the horizontal axis, following the polar coordinate convention where angles are measured counterclockwise from the horizontal. After fine-tuning parameters, the method successfully identified roofline orientations consistent with the architectural structure visible in the image. Complementary Harris corner detection further localized window intersections, and morphological segmentation refined window regions, visually confirmed by red overlays. This multi-step pipeline demonstrates a robust, noise-resilient approach for architectural feature extraction, particularly in complex natural scenes lacking ground-truth annotations, thereby contributing valuable methodologies for automated structural analysis in remote sensing and computer vision applications \cite{ref55, ref56}.

\begin{figure}[ht]
    \centering
    \subfloat[Preprocessed image.]{
        \includegraphics[width=1\linewidth]{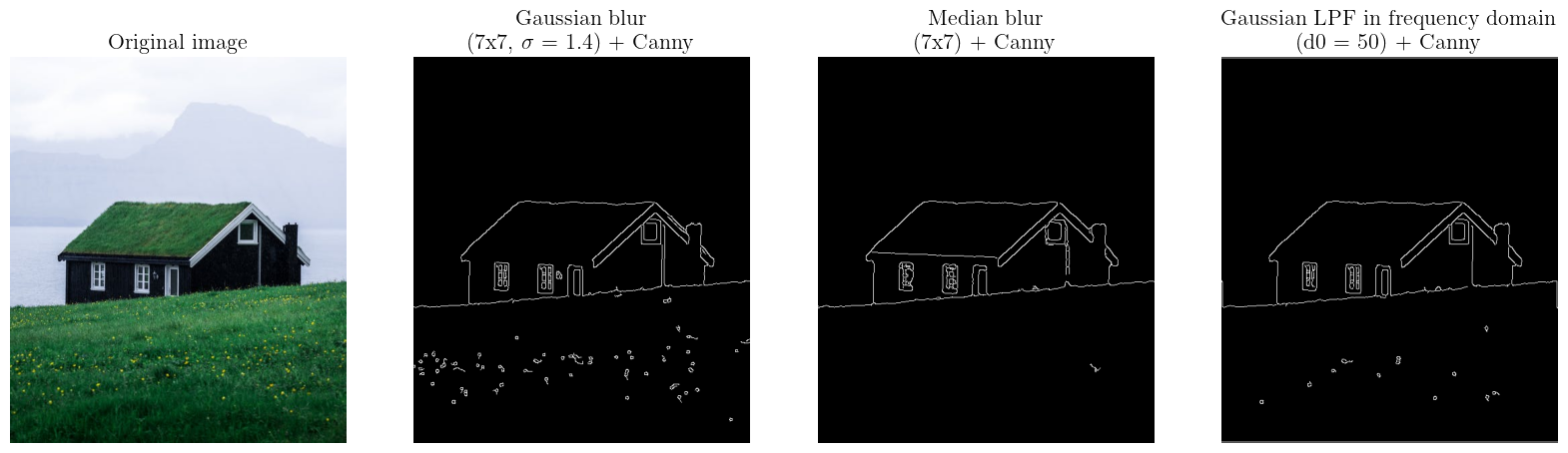}
        \label{fig:6a}
    }\\[0.5em]
    \subfloat[Canny edge detection result.]{
        \includegraphics[width=1\linewidth]{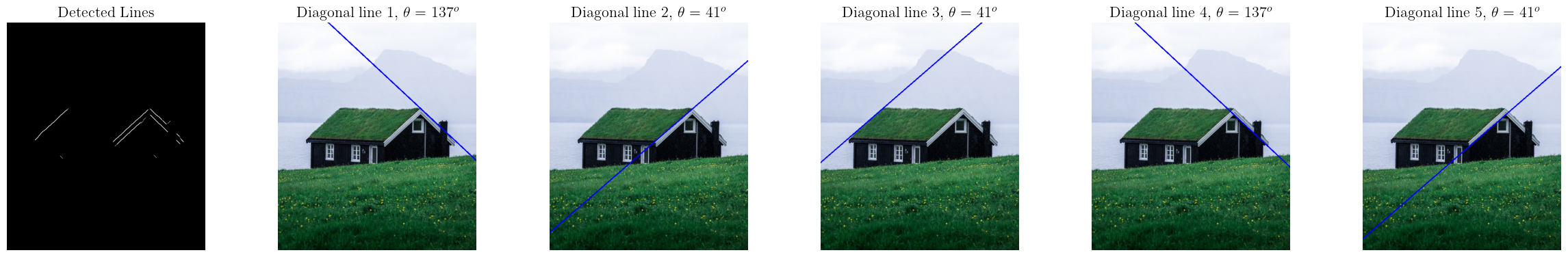}
        \label{fig:6b}
    }\\[0.5em]
    \subfloat[Line extraction using Hough transform.]{
        \includegraphics[width=1\linewidth]{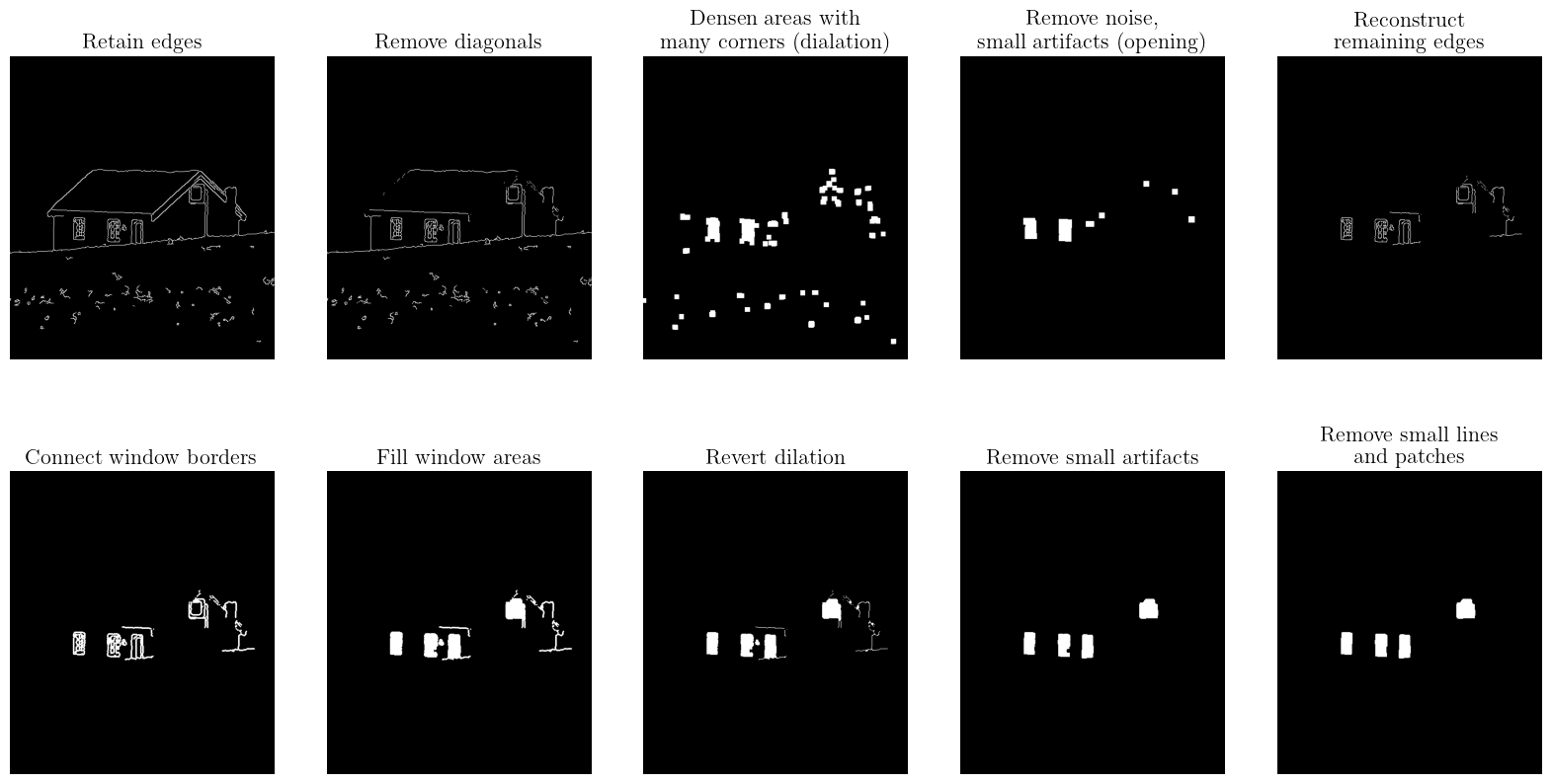}
        \label{fig:6d-1}
    }
    \caption{Edge detection and line extraction: comparison of preprocessing techniques with the Canny operator.}
    \label{fig:reconstruction-stages}
\end{figure}

We applied Harris corner detection to \texttt[image11]'s third phase of analysis, edge detection, and line extraction. The image was first converted to grayscale and filtered using a median filter to reduce noise, especially from the grassy areas that could cause false corner detections. This preprocessing ensured that detected corners corresponded mainly to meaningful edge intersections. Building upon these corners, morphological operations were employed to isolate window regions. Edges obtained via a 5 x 5 median filter underwent morphological openings with diagonal and anti-diagonal kernels to remove irrelevant diagonal edges. Subsequently, dilation and opening consolidated clustered corners into contiguous areas, enabling edge reconstruction around these regions \cite{ref57, ref58}. Further dilation and hole-filling created continuous window surfaces, refined by erosion and morphological filtering to eliminate noise and small artifacts. Finally, connected component analysis removed rectangular shapes presumed to be doors based on spatial context. This integrated approach demonstrated effective architectural feature extraction in complex backgrounds.

\begin{figure}[ht]
    \centering
    \subfloat[Window Localization]{%
        \includegraphics[width=0.48\linewidth]{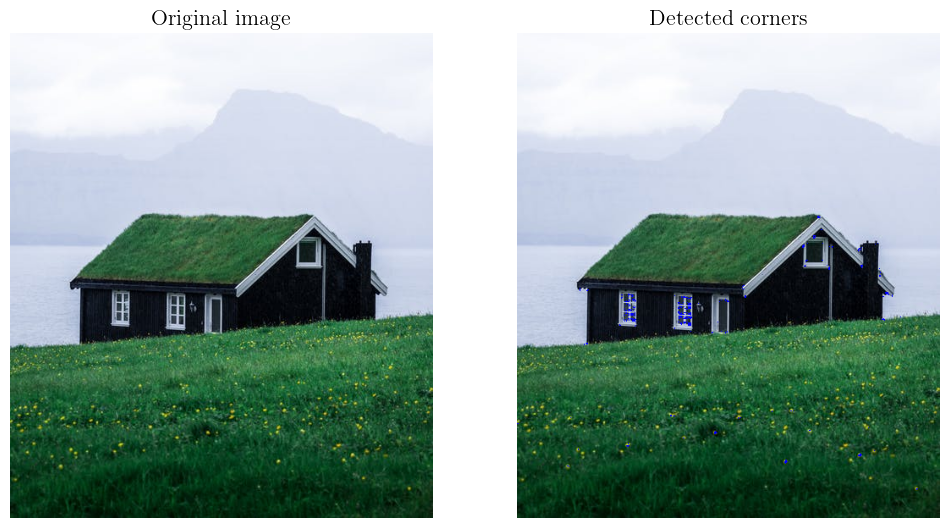}%
        \label{fig:6c}%
    }
    \hfill
    \subfloat[Corner Extraction]{%
        \includegraphics[width=0.48\linewidth]{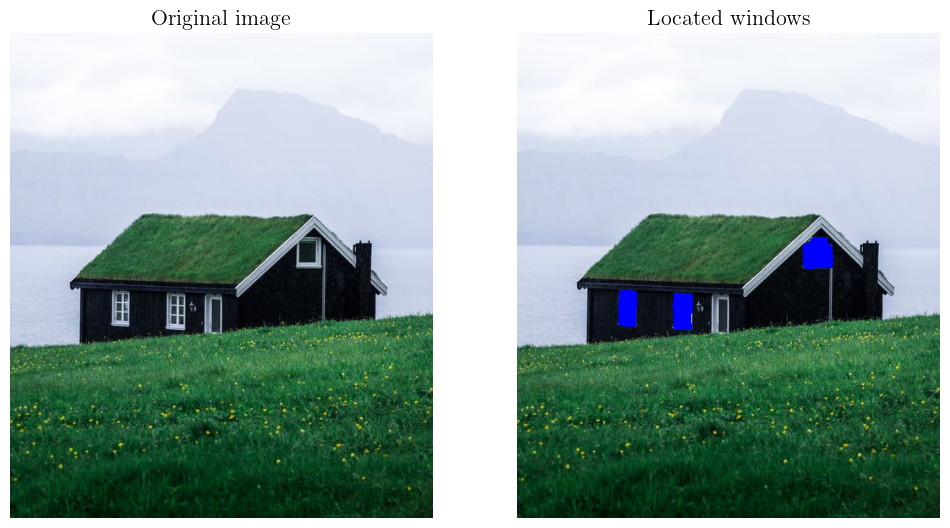}%
        \label{fig:6d-2}%
    }
    \caption{Feature detection pipeline: (a) window localization, (b) corner extraction.}
    \label{fig:side-by-side-6}
\end{figure}

\subsubsection{Cue Isolation and Rotation}
For \texttt{image31}, Hough-based angle estimation produced an orientation of 51.50$^\circ$, which, upon rotation, aligned the cue with \texttt{image32} and yielded an 86.07\% blended similarity score. Furthermore, cue isolation achieved an 81.87\% similarity with \texttt{image33} through thresholding combined with morphological processing. The quantitative outcomes are summarized in Table~\ref{tab:combined_scores}. Despite minor artifacts introduced by ball occlusion, visual inspection confirms that the proposed method ensures accurate cue alignment and isolation.

\begin{algorithm} [ht]
\caption{Billiard Cue Isolation}
\label{alg:cue_isolation}
\begin{algorithmic}[1]
\Function{CueIsolation}{$I_{\text{RGB}}$}
    \State $E \gets$ CannyEdge($I_{\text{RGB}}$, 100, 200)
    \State $L \gets$ HoughLines($E$, $\rho=1$, $\theta=1^\circ$, thresh=200)
    \State $\theta_{\text{avg}} \gets$ mean($\theta$ from $L$)
    \State $\theta_{\text{deg}} \gets 90 - (180 \cdot \theta_{\text{avg}} / \pi)$
    \State $B \gets$ HoughCircles($I_{\text{RGB}}$, $r \in [25, 33]$) \Comment{Detect balls}
    \State $I' \gets$ fillCircles($I_{\text{RGB}}$, $B$, color=black)
    \State $H \gets$ histogram($I'$, bins=256)
    \State $I'' \gets$ threshold($I'$, $H_{\text{peak}} = 49$, value=black)
    \State $E'' \gets$ HoughLines($I''$, $\rho=1$, $\theta=1^\circ$, thresh=200)
    \State $I''' \gets$ maskNonCue($I''$, $E''$, value=black)
    \State $I_{\text{final}} \gets$ rotate($I'''$, $-\theta_{\text{deg}}$)
    \State \Return $I_{\text{final}}$
\EndFunction
\end{algorithmic}
\end{algorithm}

Algorithm~\ref{alg:cue_isolation} isolates the billiard cue from an RGB image by combining edge detection, line detection, and geometric masking techniques. First, edges are extracted using the Canny detector, and prominent lines are identified via the Hough transform to estimate the cue's average orientation. Simultaneously, billiard balls are detected with circular Hough transforms and masked to prevent interference. The image is then thresholded based on the histogram peak, and remaining non-cue elements are removed using line-based masking. Finally, the image is rotated to align the cue horizontally, producing a final image $I_{\text{final}}$ containing only the isolated billiard cue. This method effectively separates the linear cue from surrounding objects and background, facilitating subsequent analysis such as trajectory tracking or automated game recognition.

\begin{figure}[ht]
    \centering
    \subfloat[]{%
        \includegraphics[width=1\linewidth]{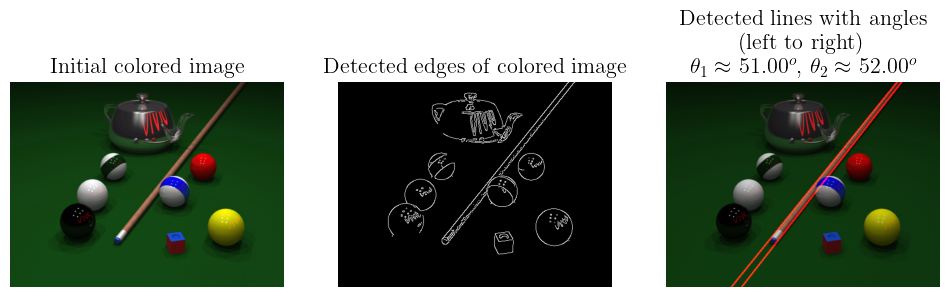}%
        \label{fig:7a-1}%
    }\\[0.5em]
    \subfloat[]{%
        \includegraphics[width=1\linewidth]{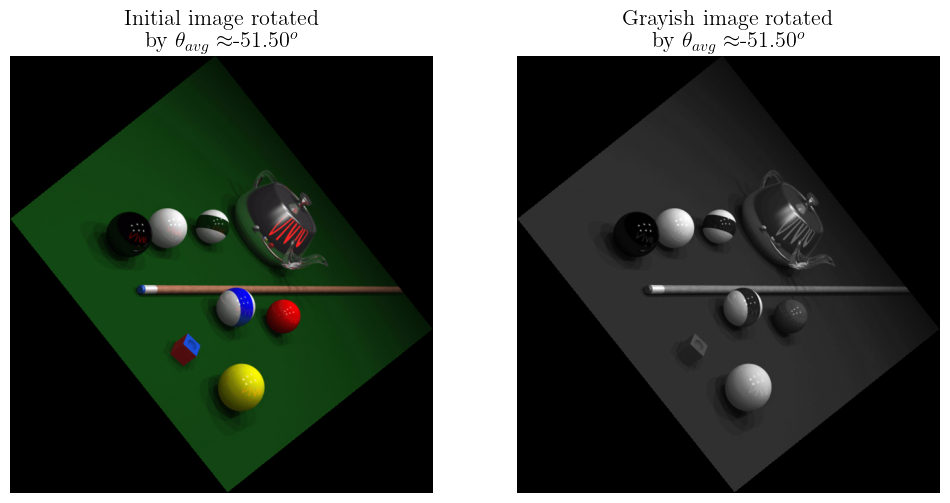}%
        \label{fig:7a-2}%
    }
    \caption{Image rotation correction using average angle estimation ($\theta_{\text{avg}} = 51.50^\circ$).}
    \label{fig:combined-7a}
\end{figure}

The angle of the lines intersecting the edges of the cue was determined by converting the BGR image to RGB color space and measuring the angles at the intersections. Edge detection was performed using the Canny algorithm with optimized hysteresis thresholds (100 and 200). Subsequently, the Hough Line Transform identified the cue’s bounding lines, with parameters tuned for line width (1 pixel) and accumulator threshold (200). Notably, the method yielded superior results on the RGB image compared to grayscale. The detected lines, defined by parameters \(\rho\) and \(\theta\), were used to calculate the average angle, converted from radians to degrees, resulting in 38.5°. Since this was complementary to the desired angle, it was adjusted to 51.5°. The image was then rotated using \texttt{ndimage.rotate} with a counterclockwise correction. Further processing isolated the cue by blackening billiard balls and background areas via circle detection and pixel intensity thresholding, yielding a rotated grayscale image with the cue successfully segmented.

\begin{figure} [ht]
    \centering
    \includegraphics[width=1\linewidth]{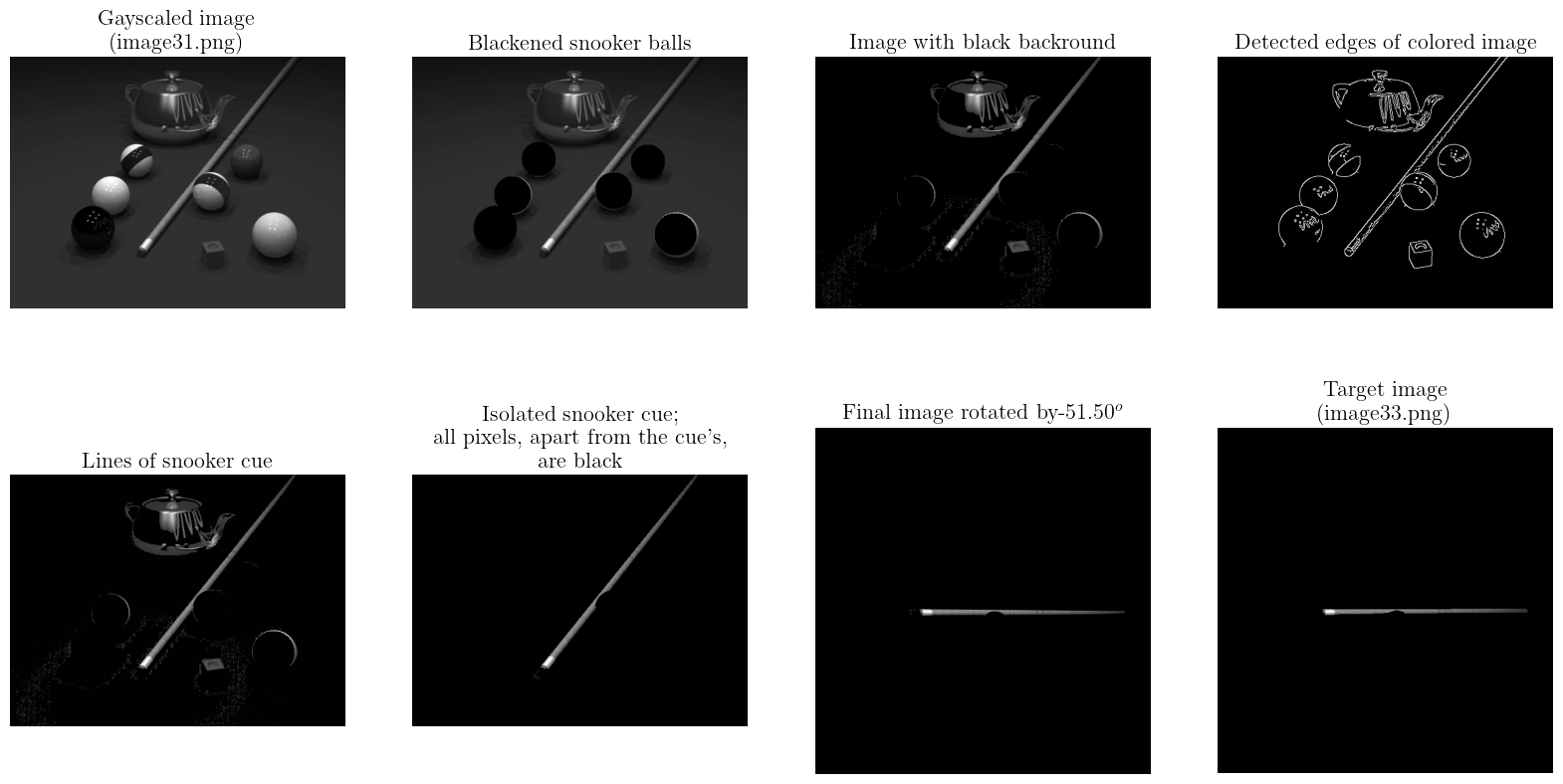}
    \caption{Snooker cue detection and alignment pipeline: (a) Grayscale conversion of input image (image31), (b) Background suppression and ball isolation, (c) Edge detection results, (d) Cue line extraction, (e) Cue isolation via binary masking, (f) Rotation correction by$-51.50^\circ$, and (g) Target reference image (image33). }
    \label{fig:enter-label}
\end{figure}

\section{Discussion}
The proposed spatial processing methods demonstrate consistent effectiveness across image analysis tasks, emphasizing precision and generalizability. Grayscale quantization using a stepwise transformation reduces intensity levels to eight, yielding a visually distinct posterization effect in \texttt{nature\_dark\_forest.jpg}, as corroborated by discrete histogram distributions. In contrast enhancement (step 2), YCrCb histogram equalization outperforms RGB-based techniques by enhancing luminance while preserving chrominance, thus avoiding hue distortion—a critical requirement in natural image contexts. Brightness enhancement via HSV space (Exercise 3) increases perceptual clarity without compromising color fidelity, demonstrated effectively in \texttt{pollen-500x430px-96dpi}. 

The custom sharpening kernel enhances high-frequency components in lunar images, accentuating surface detail with minimal artifact introduction \cite{ref48}. Bidirectional pipeline using spatial transforms achieves a 76.10\% forward and 74.80\% reverse similarity (blended SSIM+NMI), validating its robustness under noise and contrast variation. Geometric feature extraction reliably identifies architectural structures, including diagonal rooflines and window corners, with Hough-based angle estimation achieving accurate localization \cite{ref49}. Finally, cue isolation and alignment (step 7) yield high similarity scores (86.07\% and 81.87\%), with a precisely computed 51.50$^\circ$ rotation angle. These results underscore the practicality of the proposed deterministic techniques for real-time and low-level vision tasks.

\subsection{Limitations and Challenges}
Despite their utility, several limitations constrain the scalability of the proposed methods. The quantization pipeline's reliance on fixed thresholds limits adaptability across scenes with differing intensity distributions. However, RGB histogram equalization introduces chromatic distortions, motivating the use of perceptually linear spaces like YCrCb or HSV. The transformation pipeline in the reverse mapping exhibits fidelity loss (74.80\%) due to amplified noise and structural degradation in the forward stage, highlighting the non-reversibility of certain nonlinear filters \cite{ref58}. Feature extraction via Canny and Hough transforms is sensitive to parameter selection, requiring manual tuning that may fail under illumination changes or textured backgrounds. Despite being accurate under ideal conditions, window detection is affected by occlusions, irregular surfaces, and architectural clutter, necessitating heavy morphological filtering. In cue isolation, ball occlusions and non-uniform table textures pose challenges, sometimes resulting in incomplete segmentation \cite{ref60}. Furthermore, the computational cost of iterative morphological and Hough-based methods limits their viability for real-time or embedded deployment. This constraint needs to be overcome with a framework that is based on adaptive learning for optimization.

\section{Conclusion}
This study presents a modular and interpretable framework for spatial image processing, evaluated across diverse low-level vision tasks. The framework integrates multiple complementary components, including grayscale quantization, color space enhancement, image sharpening, geometric feature extraction, and object segmentation. Stepwise quantization reduces intensity levels while preserving structural details, whereas YCrCb histogram equalization and HSV-based brightness adjustment improve contrast and perceptual quality without introducing chromatic distortion. High-frequency sharpening is employed to enhance fine edge details, particularly in scientific imagery. A bidirectional transformation pipeline is proposed to model reversible contrast and noise variations, achieving similarity scores of 76.10\% (forward) and 74.80\% (reverse). Geometric approaches—including Canny, Hough, and Harris-based methods—enable accurate architectural feature extraction and robust cue segmentation, with alignment performance exceeding 86\% similarity. Evaluation of a blended SSIM + NMI metric, ensuring both perceptual and statistical fidelity. While the framework is efficient and deterministic, its adaptability remains limited under varying noise conditions. Future research will focus on adaptive thresholding, learning-based filtering, and transformer-driven enhancement to improve robustness and generalization. Moreover, replacing rule-based pipelines with deep segmentation architectures (e.g., U-Net, Mask R-CNN) is recommended, particularly beyond static imagery. The framework can be extended to video, multispectral, and 3D vision tasks, opening applications in autonomous systems, remote sensing, and biomedical imaging.

\end{document}